\documentclass[letterpaper, 10 pt, conference]{ieeeconf}  

\IEEEoverridecommandlockouts                              

\usepackage{amssymb}  

\usepackage{graphicx}
\usepackage{color,soul}
\usepackage[caption=false,font=footnotesize]{subfig}
\usepackage{booktabs}
\usepackage{changes}
\usepackage{amsmath, amssymb}         
\usepackage[ruled,linesnumbered]{algorithm2e}  
\usepackage{enumerate}

\usepackage{threeparttable}

\usepackage[normalem]{ulem}
\def\modified_color{black}
\def\rev_color{black}

\newcommand{\bm}[1]{\mbox{\boldmath $ #1 $}}
\newcommand{\figref}[1]{{Fig. \ref{#1}}}

\newcommand{\sectref}[1]{{Section \ref{#1}}}

\usepackage[T1]{fontenc}
\usepackage{newtxtext,newtxmath} 

\usepackage[unicode]{hyperref}

\title{\LARGE \bf
CoTaP: Compliant Task Pipeline and Reinforcement Learning of Its Controller with Compliance Modulation
}

\author{Zewen He, Chenyuan Chen, Dilshod Azizov, Yoshihiko Nakamura
\thanks{The authors are with the Department of Robotics,
Mohamed bin Zayed University of Artificial Intelligence, Masdar City, Abu Dhabi, United Arab Emirates. 
        {\tt\small zewen.he@mbzuai.ac.ae}}%
}

\begin{document}

\maketitle
\thispagestyle{empty}
\pagestyle{empty}

\begin{abstract}
Humanoid whole-body locomotion control is a critical approach for humanoid robots to leverage their inherent advantages. 
Learning-based control methods derived from retargeted human motion data provide an effective means of addressing this issue. 
However, because most current human datasets lack measured force data, and learning-based robot control is largely position-based, achieving appropriate compliance during interaction with real environments remains challenging. 
This paper presents Compliant Task Pipeline (CoTaP): a pipeline that leverages compliance information in the learning-based structure of humanoid robots. 
A two-stage dual-agent reinforcement learning framework combined with model-based compliance control for humanoid robots is proposed. 
In the training process, first a base policy with a position-based controller is trained; then in the distillation, the upper-body policy is combined with model-based compliance control, and the lower-body agent is guided by the base policy. 
In the upper-body control, adjustable task-space compliance can be specified and integrated with other controllers through compliance modulation on the symmetric positive definite (SPD) manifold, ensuring system stability. 
We validated the feasibility of the proposed strategy in simulation, primarily comparing the responses to external disturbances under different compliance settings. 
For detailed experimental results, please see the attached video. \footnotemark
\end{abstract}

\footnotetext{Attached video: \url{https://drive.google.com/file/d/1Ge08DPEVZRw04pIZNqqBSJQaBJkdImoH/view?usp=sharing}}

\section{Introduction\label{sect:introduction}}

In recent decades, humanoid robot technology has made significant advancements. 
Particularly over the past five years, there has been a surge in the development of diverse humanoid robot body designs, including Atlas from Boston Dynamics, Optimus from Tesla, Figure’s humanoid robots, and Unitree's humanoid robots like H1 and G1. 
Meanwhile, with the rapid progress in the field of artificial intelligence, reinforcement learning (RL) and imitation learning (IL) have been increasingly applied to the control of humanoid robots, leading to significant breakthroughs. 
In the first step, the imitation motion controller was applied in the simulation for humanoid character control, such as in \cite{peng2018deepmimic,peng2021amp}. 
After that, this approach was extended into real humanoid robot whoel-body control (WBC) \cite{tang2024humanmimic,zhang2024whole}. 
Based on the learning method, the controller no longer requires accurate modeling of the robot and environment such as model predictive control (MPC), which demonstrates improved robustness and generalizability in complex environments.

Humanoid robots have the key features that can simultaneously perform locomotion and manipulation, which is abbreviated as \emph{loco-manipulation} \cite{gu2025humanoid}. 
In many studies, the upper and lower bodies of humanoid robots are controlled independently. During manipulation tasks, the legs are typically kept stationary to maintain balance and ensure stability of the center of mass (CoM) \cite{rakita2019shared}.
In recent studies, IL based on human motion data has also been applied to loco-manipulation. The data sources include publicly available human motion datasets as well as data collected through tele-operation \cite{he2024learning,fu2024humanplus}.

\begin{figure}[t]
  \centering
  \subfloat[\label{fig01a}]{
    \includegraphics[width=.3\linewidth]{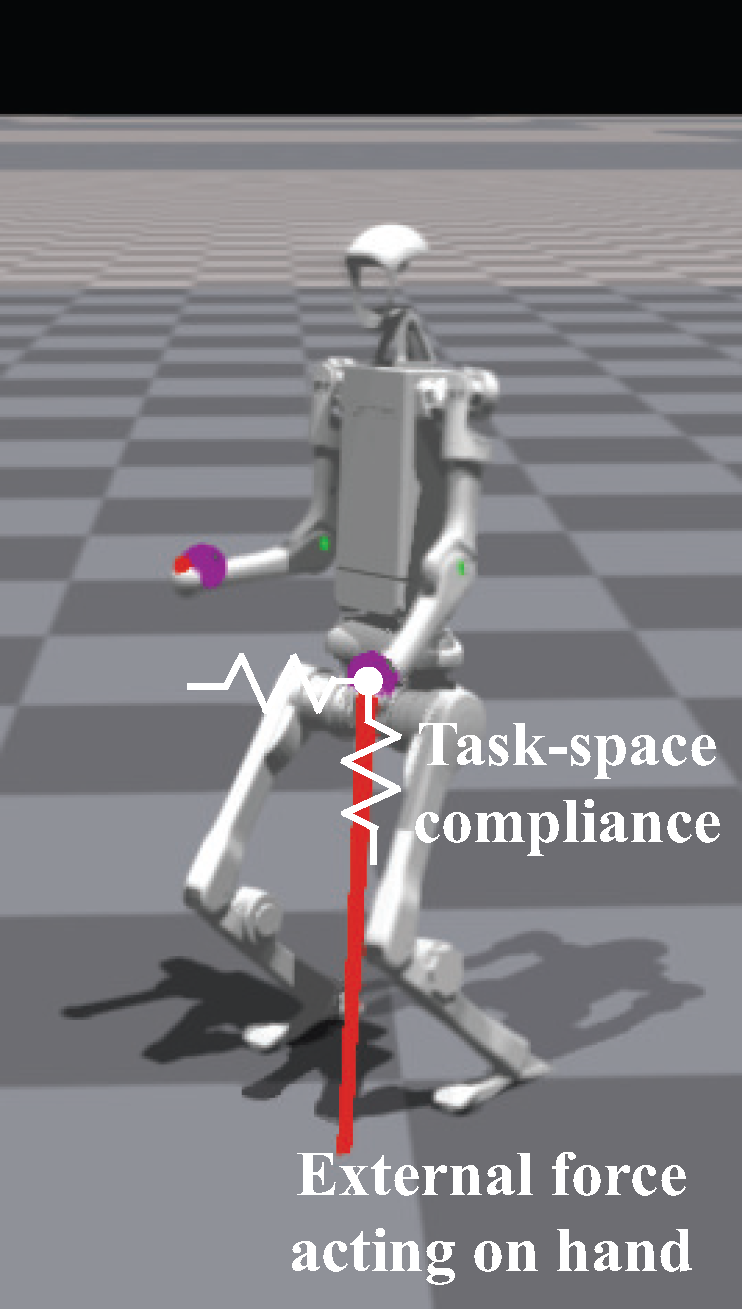}
  }\hfill
  \subfloat[\label{fig01b}]{
    \includegraphics[width=.3\linewidth]{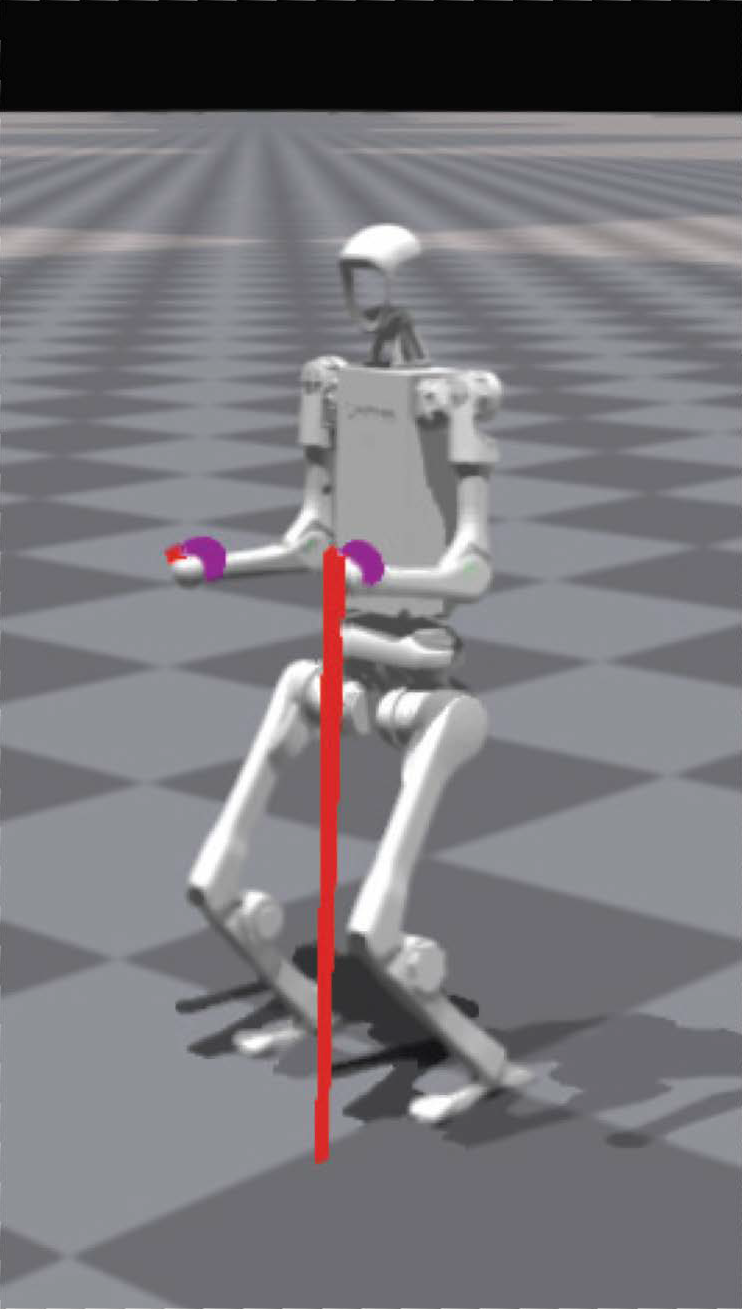}
  }\hfill
  \subfloat[\label{fig01c}]{
    \includegraphics[width=.3\linewidth]{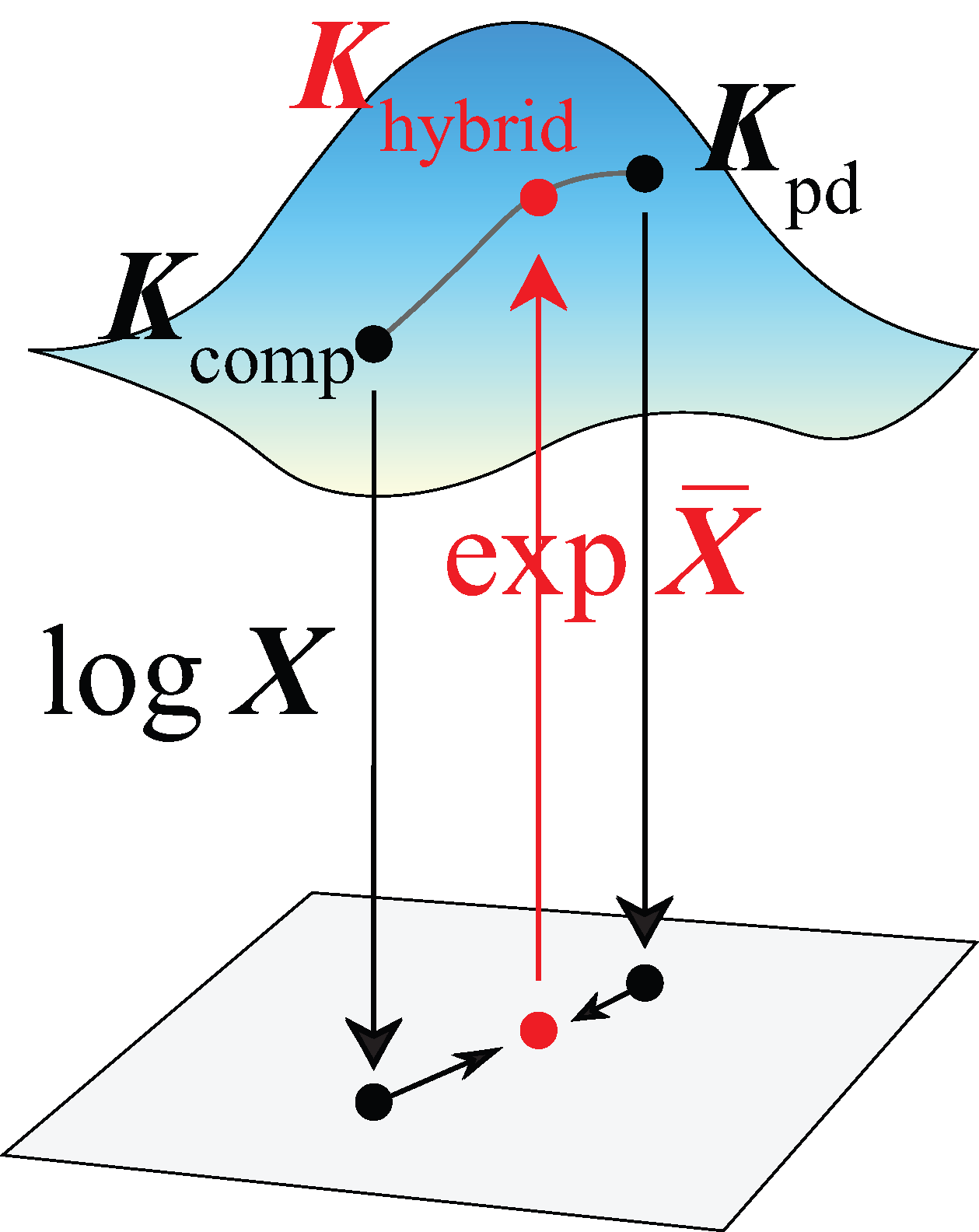}
  }
  \caption{(a) Simulation of H1 under a vertical load in the low-stiffness condition. (b) Simulation of H1 under a vertical load in the high-stiffness condition. (c) Illustration of the stiffness matrices modulation on SPD manifold. Two different original stiffness matrices are first mapped to the Log-Euclidean space using the log mapping, then linearly interpolated, and finally mapped back using the exp mapping. }
  \label{fig:init}
\end{figure}

There are several key aspects in the current research receiving significant attention. 
First, most current research focuses on joint space PD control for humanoid motion control. 
This control approach may yield satisfactory results in current purely motion control scenarios; however, it fails to implement force control when interactions with the environment (such as manipulation and multi-contact motion) or even human-robot interactions (HRI) occur, thereby making it difficult to achieve desirable outcomes. 
Then, most human even humanoid robot data are only proprioception-based, which lacks sensory input and action output. 
Although some studies have already attempted to incorporate contact force data into human motion data collection \cite{pham2017hand,pham2017multicontact}, the overall size and generality of such datasets remain insufficient. 
In addition, the commonly used physical simulators make sim-to-real transfer more difficult, as they lack accuracy and are computationally expensive when simulating complex contacts. 
Therefore, a key challenge lies in how to leverage the currently limited data resources to achieve force control for humanoid robot loco-manipulation.

To address these challenges in a comprehensive manner, we turn to a classical topic in traditional robotics: compliance control. 
Compliance control, by adopting a relatively passive mechanism, is capable of adapting to unknown or inaccurate contacts. Such a property is exactly what is required to overcome the training challenges arising from the lack of sufficient contact information;
moreover, compliance control inherently provides force control capabilities, which makes it particularly suitable for addressing the challenges of real-world robot–environment interactions in loco-manipulation tasks. 
In model-based robot control theory, compliance control is a practical method to guarantee robot interaction safety and stability with the environment. 
\cite{henze2015approach,dean2019whole} proposed the hierarchical compliance control method for humanoid robot. 
In the study \cite{Yamamoto2017ICRA,yamamoto2021humanoid}, the authors optimized the joint-space viscoelasticity matrices by an analytical way. 
This work has enabled compliance control to achieve promising results in maintaining balance in humanoid robots. 
In the field of robotic manipulation, compliance control is even more critical. It plays a key role in ensuring the robot's safety, enhancing HRI, and enabling adaptive manipulation capabilities \cite{khan2014compliance,ott2015prioritized}. 

\begin{figure}[t]
	\begin{center}
    \centering
	 \includegraphics[width=0.99\hsize]{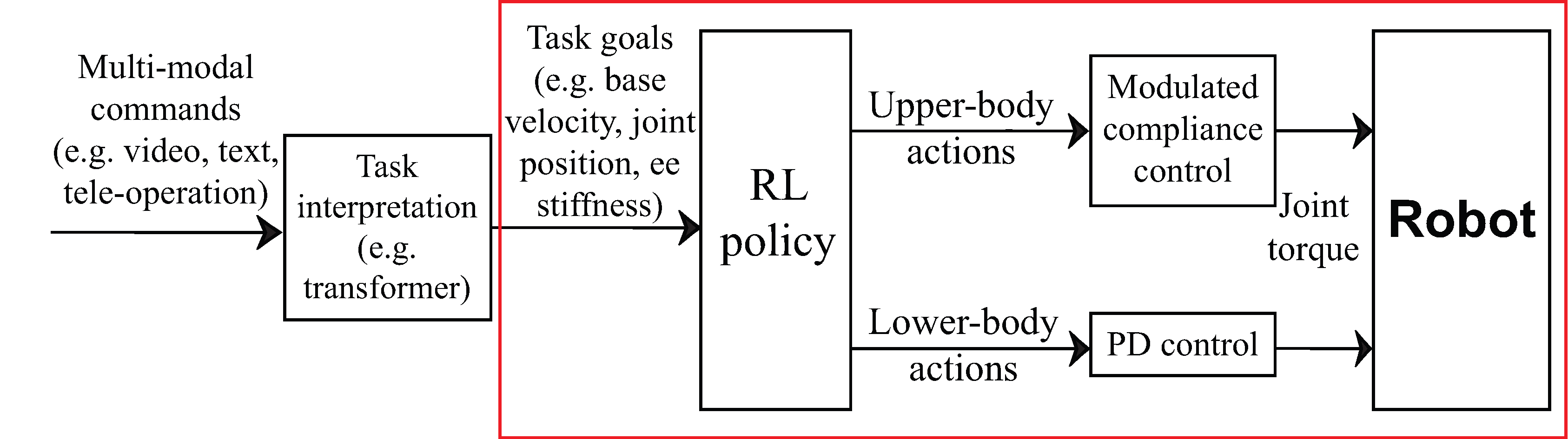}
	 \caption{Overview of CoTaP pipeline. The red frame highlights the method proposed in this paper, and our objective is to implement the entire pipeline on humanoid robots. }
    \label{fig:cotap}
	\end{center}
\end{figure}

Compared with model-based control, the greatest advantage of RL lies in its robustness and generalization in complex and dynamic tasks. 
On this basis, RL method has also been integrated into compliance control. 
In \cite{lee2022deep}, the controller achieves compliant behavior by modulating the target position, which is fundamentally akin to admittance control rather than direct force control.
In other related studies \cite{hartmann2024deep,spoljaric2025variable,watanabe2025learning}, the RL-based controller typically focuses on optimizing the PD gains of individual joints or a single limb, without considering whole-body compliance. 
Recently, some methods based on model-free RL for compliance control and even force control have been proposed \cite{xu2025facet,zhang2025falcon,wei2025hmc}, but their accuracy has not been guaranteed.

Accordingly, the immediate task is to establish how parameter adjustability and stability of compliance control can be ensured within the RL framework, and to further verify the compliance effect under external perturbations. 
Therefore, we propose \textbf{Co}mpliant \textbf{Ta}sk \textbf{P}ipeline (CoTaP), a pipeline leveraging the compliance information in the humanoid robot loco-manipulation control. 
As illustrated in \figref{fig:cotap}, in this paper we mainly focus on the compliance modulation on the RL framework of the pipeline. 
In this study, we present a compliance control approach that integrates the RL control framework with the robot’s kinematic model. By performing stiffness matrix modulation on the symmetric positive definite (SPD) manifold, the stability of the joint-space control is guaranteed. 
For humanoid whole-body control, a two-stage dual-agent policy training framework is applied in our work. 

The main contributions of this study are as follows: 
\begin{itemize}
    \item[1.] Combined model-based compliance control with humanoid robot reinforcement learning control framework, designing a learning-based compliance control strategy including dual-agent policy and compliance modulation on SPD manifold for upper-body control; 
    \item[2.] Validated effectiveness of the proposed compliance control, achieving compliance modulation performance of a humanoid robot in simulation. 
\end{itemize}
\section{Related Works\label{sect:background}}




\subsection{Reinforcement Learning Based Humanoid Locomotion Control}

In recent years, there has been a surge of research on humanoid robot control that leverages human motion data retargeting in combination with deep reinforcement learning and imitation learning. 
The main prevailing trends include the use of adversarial learning such as GAIL \cite{ho2016generative} and AMP \cite{peng2021amp}, or directly applying imitation learning to the reference motion trajectories \cite{peng2018deepmimic,he2024learning,cheng2024expressive}. 
Moreover, such imitation learning methods have been extended to enable whole-body tele-operation in humanoid robots \cite{he2024learning,ze2025twist}. 
In addition, for humanoid robot loco-manipulation, the dual-agent system with separated upper and lower body control has also achieved good results in researches \cite{zhang2025falcon,ding2025jaeger}.

\subsection{Humanoid Whole-body Compliance Control}

According to the model-based compliance control method, there is a feedback controller in the task space: 
\begin{align}
    \bm{f} &= \bm{K} \varDelta \bm{p} + \bm{D} \varDelta \bm{v}
\end{align}
Correspondingly, the joint-space feedback controller can be expressed as follows: 
\begin{align}
    \bm{\tau} &= \bm \tau_{grav} + \bm{K}_{\theta} \varDelta \bm{\theta} + \bm{D}_{\theta} \varDelta \dot{\bm{\theta}} \label{eq:joint_pd}
\end{align}
where $\bm \tau_{grav}$ is the gravity compensation term. 
Under this setting, especially for redundant robots, a key research challenge in compliance control lies in establishing the mapping between task space and joint space while simultaneously considering null-space control. 
Studies such as \cite{sentis2010compliant,Yamamoto2017ICRA,yamamoto2021humanoid} have addressed this issue from both kinematic and dynamic perspectives. 
It should also be noted that compliance and stiffness form a dual relationship; therefore, in this paper, we use the two terms interchangeably in derivations depending on the context.

\subsection{Comparison with Related Works}

Drawing on existing studies, we provide a comparison in Table \ref{tab:0}. 
We defined three dimensions for comparison: adjustable, model-aware, and stability-accounted. 
Adjustable denotes the capability to adapt task parameters online without the need for retraining; model-aware denotes whether control explicitly incorporates the robot’s kinematic or dynamic model; and stability-accounted refers to whether the stability of the control method is taken into account or formally proven. 

As shown in the table, the approach proposed in this work fulfills all three criteria. 
We provide a detailed comparison here with several studies that share objectives closely aligned with ours. 
First, FACET \cite{xu2025facet} achieves force (also impedance) control by using a simplified task-space model together with model-free RL-based tracking control, which is similar to an admittance control approach. 
However, unlike traditional robotic manipulators, current quadruped and humanoid robots find it difficult to achieve precise position tracking in the world frame using RL with low-level motor control, and consequently, the final force control objective is also affected. 
On the other hand, HMC \cite{wei2025hmc} performs weighted linear interpolation of different model-based control laws at the torque level, while the control objectives and inputs of each controller are also different. However, due to the heterogeneity of these meta-controllers, it is hard to assess the stability of the control during blending at torque level, making it difficult to guarantee that the controller will not diverge at any given moment. 
Besides, the null-space stiffness in its impedance control was not considered. 

To address the limitations of the above-mentioned studies, this paper proposes a method that integrates model-based control within the RL framework to achieve compliance control with guaranteed stability.


\begin{table}[t]
\centering
\caption{Comparison between different learning-based compliance control method \label{tab:0}}
\begin{threeparttable}
\begin{tabular}{cccc}
\hline
\textbf{Method} & \textbf{Adjustable} & \textbf{Model-aware}\tnote{2} & \textbf{Stability-accounted}\\
\hline
DCC  \cite{lee2022deep}   & No & Yes & No \\
FALCON \tnote{1}    & No & No & No \\
FACET   & Yes & No & No \\
HMC    & Yes & Yes & No \\
Ours      & Yes & Yes & Yes \\
\hline
\end{tabular}
\begin{tablenotes}
\footnotesize
\item[1] FALCON achieves adaptive force control, but can be considered as passive compliance. 
\item[2] Model means explicit robot kinematic or dynamic model in controller. 
\end{tablenotes}
\end{threeparttable}
\end{table}

\section{Task-space Compliance Control in Reinforcement Learning Structure \label{sect:compliance}}

\begin{figure*}[t]
	\begin{center}
	 \includegraphics[width=0.98\hsize]{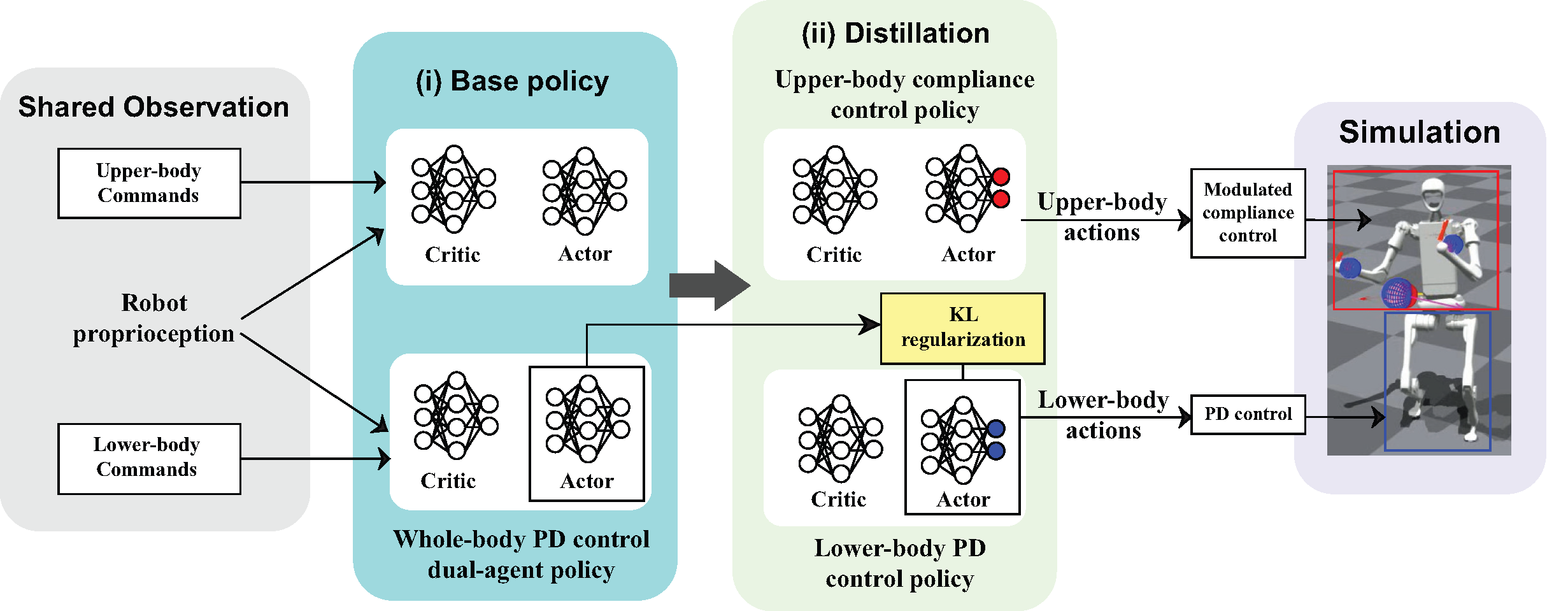}
	 \caption{Overview of the training framework in this study. }
    \label{fig:overview}
	\end{center}
\end{figure*}

\subsection{Dual-agent Learning Strategy}

In this paper, we take \cite{zhang2025falcon} as the primary baseline, and the basic learning network settings are largely based on this study. 
Similar as \cite{zhang2025falcon,ding2025jaeger,wei2025hmc}, we divide the whole body into two parts: upper- and lower-body. 
The torso is defined as the base link, which simultaneously functions as the separation link between the upper and lower body. 
Each part shares the same observation but have separate policies and actions. 
\figref{fig:overview} shows the overview of the training framework of our work. 
The actions are the whole-body target joint angle $\bm a_t = \bm q^{*}$ ($*$ means reference). 
The robot state is defined as 
$\bm s_t := [\bm q_{t-4:t}, \dot{\bm q}_{t-4:t}, \bm \omega^{\text{torso}}_{t-4:t}, \bm g_{t-4:t}, \bm \tau^{\text{upper}}_{t-4:t}, \bm a_{t-5:t-1}]$,
which contains five-step histories of joint positions, joint velocities, root angular velocity, projected gravity, upper-joint control torques and previous actions. 
The goal space $\mathcal{G}_t = [\mathcal{G}_t^l, \mathcal{G}_t^u]$ consists of locomotion goals
$\mathcal{G}_t^l := [\bm v_t^{\text{torso}*}, h_t^{\text{torso}*}, w_t^{\text{yaw}*}]$,
specifying desired torso linear velocities, torso heights, and torso yaw angles, and manipulation goals
$\mathcal{G}_t^u := [\bm q_t^{\text{upper}*}, \bm K^{\text{ee}*}_t, k^{\text{null}*}_t, \alpha]$,
specifying target joint configurations for the upper body, target task- and null-space stiffness of end-effector ($ee$ means end-effector), and compliance modulation ratio.

For the lower body, we can set target position and applying impedance control to obtain the required velocity as in \cite{xu2025facet} for tracking. 
In this study, we set the target torso position same as in reference motion datasets, and apply impedance control to get the required velocity for next step. 
In the upper body, we apply task-space compliance control to obtain a stiffness matrix. In the lower body, a simple PD control is applied.

\subsection{Decoupled Upper-body Compliance \label{sect:upper_compliance}}

For upper-body compliance control, we consider the arms are based on the torso link, therefore the upper-body compliance is directly related to the torso-link compliance. 
The joint-space stiffness matrix including upper and torso can be expressed as block matrices:
\begin{align}
    \bm K_q := \begin{bmatrix}
        \bm K_{torso} & \bm O \\
        \bm O & \bm K_u
    \end{bmatrix}
\end{align}

According to virtual work principle in quasi-static assumption, we can derive the compliance relationship between joint and task space (end-effector in the world frame): 
\begin{align}
\label{eq:compliance_rela}
    \bm C_e = \bm J_e \bm K_q^{-1} \bm J_e^{\top}
\end{align}
where $\bm J_e$ is the Jacobian matrix of the end-effector velocity. 
Applying the block division on the Jacobian matrix $\bm J_e = \begin{bmatrix}
        \bm J_{eb} & \bm J_{eu}
    \end{bmatrix}$,
we have
\begin{align}
    \bm C_e = \bm J_{eb} \bm K_{torso}^{-1} \bm J_{eb}^{\top} + \bm J_{eu} \bm K_u^{-1} \bm J_{eu}^{\top}
\end{align}
According to \cite{Yamamoto2017ICRA}, the solution of upper-body joint compliance matrix is
\begin{align}
    \bm K_u^{-1} = \bm J_{eu}^{\sharp} \widehat{\bm C}_e \bm J_{eu}^{\sharp\top} + \bm Y - \bm J_{eu}^{\sharp} \bm J_{eu} \bm Y \bm J_{eu}^{\top} \bm J_{eu}^{\sharp\top}
\end{align}
where $\widehat{\bm C}_e := \bm C_e - \bm J_{eb} \bm K_{torso}^{-1} \bm J_{eb}^{\top}$, null-space compliance $\bm Y:= 1/k^{\text{null}}\bm I$. 

For the torso-link stiffness, due to the lower-body is totally PD-based, we can use kinematic relationship as in \eqref{eq:compliance_rela} to calculate. In this case, it is necessary to distinguish the different support case of the humanoid, such as single-support and double-support (and in rare cases, flight case). 
Nonetheless, for controller simplification, the torso stiffness can be treated as constant. 
Upper-body gravity compensation is also considered in our compliance control, which is expressed as $\bm \tau^{\text{upper}}_{grav}$.

\subsection{Compliance Modulation on Symmetric Positive Definite (SPD) Manifold}

For different control task, different compliance performance is required. 
In this study, we set resolved compliance and simple PD control as example, using a ratio variable $\alpha$ to balance the two controllers. 
Rather than simply applying scaled summation, we adopt proportional combination of the two (or more) stiffness matrices on the SPD manifold, which also named as Log-Euclidean Interpolation \cite{arsigny2005fast}: 
\begin{align}
\label{spd_modulate}
    \bm K_{\text{modulated}}^{\text{upper}} = \exp \left(
    \alpha \log \bm K_{\text{comp}}^{\text{upper}} + (1-\alpha)\log \bm K_{\text{pd}}^{\text{upper}}
\right)
\end{align}
where $\bm K_{\text{comp}}^{\text{upper}}$ is the compliance control stiffness matrix $\bm K_u$ obtained in \sectref{sect:upper_compliance}; $\bm K_{\text{pd}}^{\text{upper}}$ is the original PD control stiffness, and $\alpha$ is the ratio to modulate the stiffness between compliance control and PD control. 
\figref{fig01c} illustrates the mapping and modulation process of the stiffness matrices. 
In contrast to linear interpolation performed in Euclidean space, Log-Euclidean interpolation offers the following advantage \cite{arsigny2005fast}: 
it guarantees that the interpolated stiffness matrix lies on the SPD manifold, thus avoiding outcomes that violate the underlying manifold geometry; 
it avoids the swelling effect that may occur in linear interpolation, which can lead to physically unreasonable results. 
The above advantages are not discussed extensively in this paper; rather, we treat this method merely as a more rational approach to modulating compliance. 
Based on this method, the obtained joint stiffness after modulation is always positive definite. According to the proof in \cite{yamamoto2021humanoid}, the stability of the system at the current time can be guaranteed. 

In our strategy, the ratio $\alpha$ can be set as command but also depends on the kinematic posture of the robot. 
To avoid solution problem at the near-singularity posture of the arms, we process the original $\alpha$ taking into account the condition number of task Jacobian matrix as follows:
\begin{align}
    \widehat{\alpha} = \frac{\alpha}{1 + \max(0, \mathtt{cond\_num} - 10)}
\end{align}
where $\mathtt{cond\_num}$ is the condition number of the upper body Jacobian matrix, and $\widehat{\alpha}$ is the processed modulation ratio. In the control based on \eqref{spd_modulate}, we replace $\alpha$ by $\widehat{\alpha}$ taking into account of the condition number. 
This method achieves an effect comparable to SR-inverse, while avoiding the need for extra handling during the computation of the pseudo-inverse matrix. 

In the training of policy, we applied domain randomization on the original ratio. 
Therefore, this value can be flexibly modulated to balance the contribution of the two (or more) control laws in joint-space stiffness. Moreover, inspired by \cite{wei2025hmc}, it may also be defined as the output of the high-level controller, enabling adaptation to specific task requirements.

\subsection{Two-stage Policy Distillation \label{sect:training}}

In the training we apply a two-stage policy distillation. 
As shown in \figref{fig:overview}, (i) first, we train a base policy as whole-body PD position controller with two agents $\pi^{\mathrm{lower}}_{base}$ and $\pi^{\mathrm{upper}}_{base}$ to satisfy lower-body commands tracking and upper-body motion imitation task. 
(ii) Then, we use the base policy to guide the training of a new policy with upper body compliance control, which corresponds to policy distillation. 
Specifically, we impose a KL regularization on the lower-body policy as in \eqref{eq:l_kl}, and apply it in the actor loss calculation \eqref{eq:l_dist}: 
\begin{align}
\label{eq:l_dist}
\mathcal{L}_{\mathrm{distill}} &= \mathcal{L}_{\mathrm{PPO}} + \mathcal{L}^{\mathrm{lower}}_{\mathrm{KL}}, 
\\
\label{eq:l_kl}
\mathcal{L}^{\mathrm{lower}}_{\mathrm{KL}}
&= \beta_{KL} \, D_{\mathrm{KL}}\!\left(
\pi^{\mathrm{lower}}_{distill}(\cdot \mid \bm s_t) \,\Vert\, \pi^{\mathrm{lower}}_{base}(\cdot \mid \bm s_t)
\right),
\end{align}
where $\beta_{KL}$ is a weighting coefficient decreasing with time. 

In both training stages, PPO \cite{schulman2017proximal} is applied to maximize the cumulative rewards. 
In the reward setting, we refer to FALCON work \cite{zhang2025falcon}. 
But beyond the basic reward terms, we add an extra reward term for reducing the gap between the policy’s output target joint angles and the original reference trajectory joint angles: $\exp(-\sigma_{\text{ref}}\| \pi^{\mathrm{upper}}_{fine}(\bm s_t) - \bm q_t^{\text{upper}*}\|_2)$. 

Since the new control goals are included, the domain randomization extends the following terms: end-effector stiffness matrix, null-space stiffness matrix, modulation ratio $\alpha$. 
The arrangement of the domain randomization is shown in Table \ref{tab:1}. 


\begin{table}[t]
\centering
\caption{The Range of Randomization Added on Default Values \label{tab:1}}
\begin{tabular}{ll}
\hline
\textbf{Term} & \textbf{Value} \\
\hline
Friction coefficient    & $\mathcal{U}(0.5,\, 1.25)$ \\
Link mass      & $\mathcal{U}(0.9,\, 1.2) \times \text{default [kg]}$ \\
Base mass      & $\mathcal{U}(-1.0,\, 3.0)\,\text{[kg]}$ \\
Control delay  & $\mathcal{U}(0,\, 20)\,\text{[ms]}$ \\
P Gains (base) & $\mathcal{U}(0.9,\, 1.1) \times \text{default [Nm/rad]}$ \\
D Gains (base) & $\mathcal{U}(0.9,\, 1.1) \times \text{default [Nms/rad]}$ \\
$K^{ee}$ Gains & $\mathcal{U}(0.5,\, 1.5) \times 300.0$ [N/m] \\
$k_{null}$ Value & $\mathcal{U}(0.6,\, 1.4) \times 40.0$ [Nm/rad] \\
$\alpha$ Value & $\mathcal{U}(0,\, 1)$ \\
\hline
\end{tabular}
\end{table}

\section{Results \label{sect:simulation}}

In this section, we describe the setup for training RL policies on the Unitree H1 humanoid robot in Isaac Gym environment. 
For upper-body motion priors, we employ the AMASS dataset \cite{mahmood2019amass} filtered by Perpetual Humanoid Control (PHC) method \cite{luo2023perpetual}. These processed datasets serve as demonstrations and references for initializing and guiding the policy.
The basic network architecture and hyperparameter settings are kept consistent with those of FALCON \cite{zhang2025falcon}, ensuring comparability and leveraging prior work on scalable humanoid locomotion control.
All experiments are conducted on a workstation running Ubuntu 22.04, equipped with an Intel Core i9-14900K CPU and an NVIDIA RTX 4080 GPU.

In the specific configuration, humanoid H1 has 19 active DOFs (upper body: 8, lower body: 11). Torso link is adopted as the dividing boundary between the upper and lower body. 
3-dimension position is set as the task space for both hands. 
In current simplified situation, we ignore the effect of lower-body configuration on torso stiffness, therefore $\bm K_{torso}^{-1}$ is considered as constant. 
The upper-body compliance controller calculated the required torque and directly sent to joint actuator; while lower-body policy directly sent target joint position to actuator. The frequency of this step is 50 Hz. 
For a feasible policy training, it requires about 20,000 iterations. 

In the following simulation results, different stiffness matrices were applied in the task-space, while null-space $k_{null}$ is set as 25. 
For lower-body, all joint PD used the default values provided by Unitree (all upper-body joint $K_p$ = 100). 
As mentioned above, our method is built on the FALCON framework; therefore, we adopt FALCON method as the baseline, namely pure joint-level PD control.


\subsection{Evaluation Criterion \label{sec:crit}}
For a numerical comparison of the baseline and the proposed CoTaP, the following evaluation criteria are established: 
\begin{enumerate}
    \item Torso velocity tracking error: 
    \begin{align}
        e^{\text{torso}} = \frac{1}{T} \sum\nolimits^T_{t=0} \Bigl\| \bm v_t^{\text{torso}*} - \bm v_t^{\text{torso}} \Bigl\|_2
    \end{align}
    \item End-effector tracking error: 
    \begin{align}
        e^{\text{ee}} = \frac{1}{T} \sum\nolimits^T_{t=0} \Bigl\| \bm p_t^{\text{ee}*} - \bm p_t^{\text{ee}} \Bigl\|_2
    \end{align}
    \item Average of upper-body joint torques: 
    \begin{align}
        J^{\text{upper}} = \frac{1}{T}\sum\nolimits^T_{t=0} \Bigl\| \bm \tau_t^{\text{upper}} \Bigl\|_2
    \end{align}
    \item Upper-body tracking error: 
    \begin{align}
        e^{\text{upper}} = \frac{1}{T} \sum\nolimits^T_{t=0} \Bigl\| \bm q_t^{\text{upper}*} - \bm q_t^{\text{upper}} \Bigl\|_2
    \end{align}
\end{enumerate}

\subsection{Simulation Results}


\subsubsection{Constant Load Test on End-effector}
\begin{figure}[t]
  \centering
  \subfloat[$\bm K^{ee}=100\,\mathrm{N/m}$ (in $z$)\label{fig03-3d}]{
    \includegraphics[width=.46\linewidth]{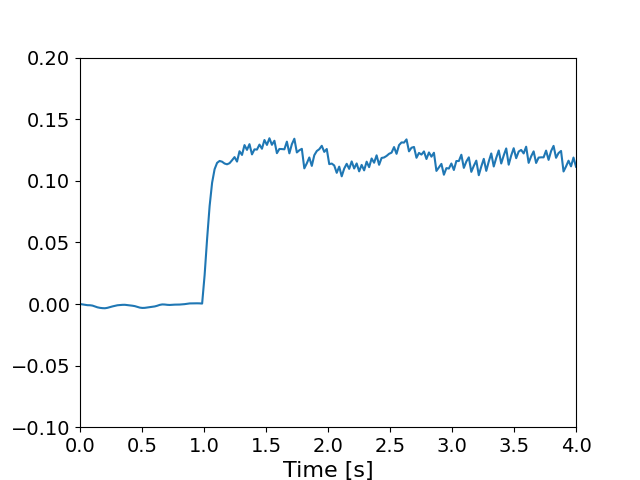}
  }\hfill
  \subfloat[$\bm K^{ee}=500\,\mathrm{N/m}$ (in $z$)\label{fig03-3a}]{
    \includegraphics[width=.46\linewidth]{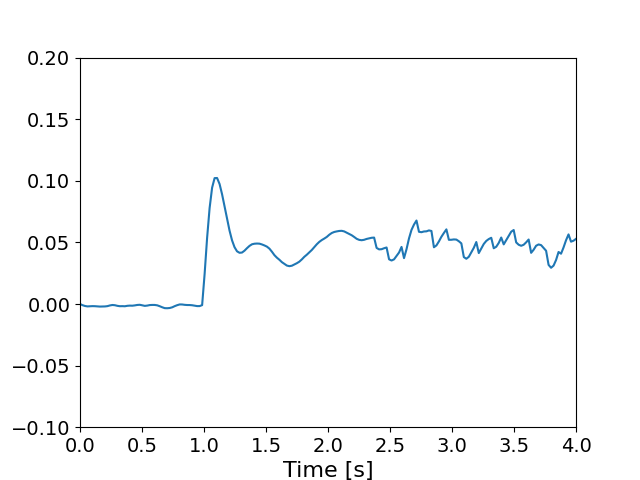}
  }\\[0.5ex]
  \subfloat[PD\label{fig03-3b}]{
    \includegraphics[width=.46\linewidth]{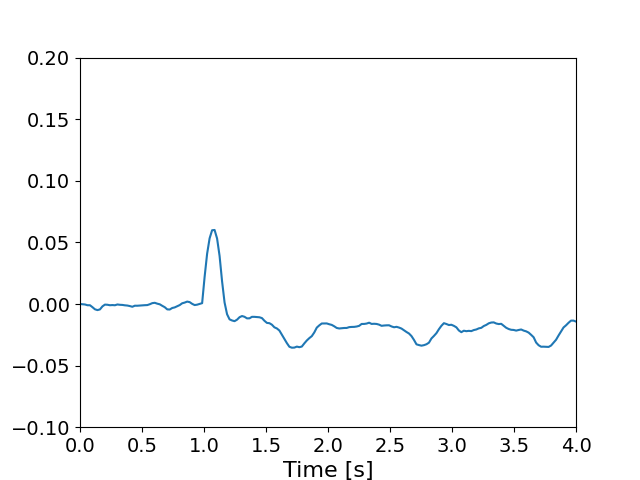}
  }\hfill
  \subfloat[$\alpha=0.5$ (PD and $100\,\mathrm{N/m}$)\label{fig03-3c}]{
    \includegraphics[width=.46\linewidth]{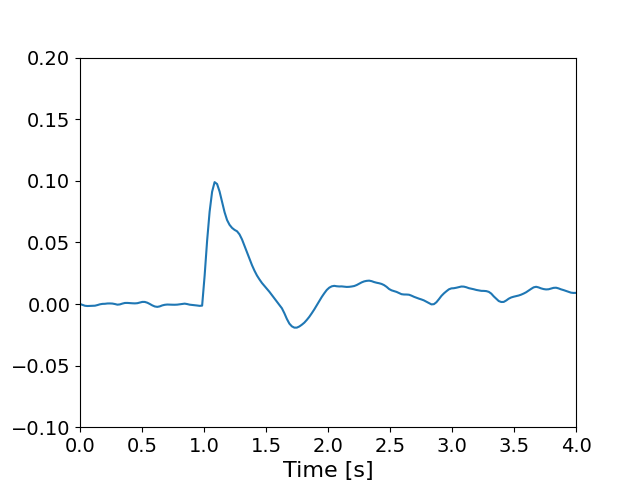}
  }
  \caption{Left hand position error in the $z$-axis under a constant $-50\,\mathrm{N}$ payload (applied after $1.0\,\mathrm{s}$). 
  Stiffness values are for the $z$-axis; $x,y$ are set to $300\,\mathrm{N/m}$. Position error unit: m.}
  \label{fig03-3}
\end{figure}

As in FALCON, an external command is employed in our controller to configure the stance and stepping (walking) modes. 
In the standing mode, the robot maintains its lower body stationary on the ground through double-support, while the arms are set to maintain their default L-shape. 
At this point, a constant vertical downward external force (-50 N in $z$-axis) is applied to the robot’s left hand to simulate the action of lifting a heavy object.
\figref{fig01a} and \figref{fig01b} are the robot in low ($\bm K^{ee}$ = 100 N/m) and high ($\bm K^{ee}$ = 1000 N/m) task-space stiffness settings, respectively. We can see the displacement of left hand are different in two cases. 
\figref{fig03-3} shows the results of hand position error in different stiffness settings of the proposed method. 
From these figures, we can observe that under small hand stiffness (i.e., high compliance, such as $\bm K^{ee}$ = 100 N/m), the hand error is relatively large and sustained (about 0.12 m); whereas under high stiffness ($\bm K^{ee}$ = 500 N/m), the hand error exhibits small oscillations and quickly decays to a constant value (about 0.05 m, respectively). 
Furthermore, as stiffness increases, the residual error at steady state decreases. 
This is consistent with the ideal task-space control objective established beforehand. 
Nevertheless, the error arises between the displacement values under different stiffness settings and the expected outcomes. 
This can primarily be attributed to two factors: (i) large displacements are restricted by the structural limits of the arm, and (ii) the RL action inherently compensates for hand errors to some extent. 
In addition, we compared the hand error when applying PD and setting $\alpha$ = 0.5 between PD and $\bm K^{ee}$ = 100 N/m. 
The results indicate that while PD control achieves a lower peak error, it subsequently exhibits reverse errors (about -0.03 m), whereas the compliance modulation method provides a favorable oscillation-damping effect. 
In other words, due to the modulation of stiffness on SPD manifold, it combines the characteristics of both control approaches, and yields improved performance.

\begin{figure*}[t]
	\begin{center}
	 \includegraphics[width=0.9\hsize]{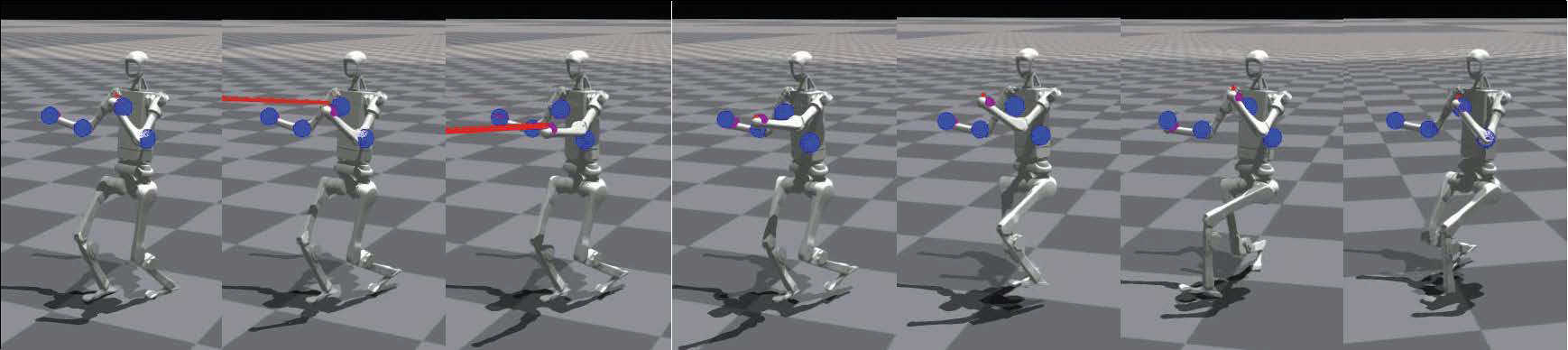}
	 \caption{Screenshots of walking control under an external impact on left hand. The upper-body reference motion is punching. The red line represents the external impact (500 N in 0.05 s). The blue balls are reference points of hands and elbows. }
    \label{fig:walk}
	\end{center}
\end{figure*}

\subsubsection{Impact Force at End-effector in Stance Mode}

In the stance mode, we added an external impact (500 N in 0.05 s, $+x$) on H1 left hand in different compliance cases. 
After the impact, the robot’s left arm was driven into large swings, while the body developed velocity errors. 
For the experimental evaluation, we measured several error metrics in \sectref{sec:crit} across 4096 environments with randomly sampled reference motions. 
The results are presented in Table \ref{table:subtab11}. 
For the proposed compliance controller (central 3 rows), task-space $\bm K^{ee}$ in $x$, $y$ axis is set as in the table, while in $z$ is 300 N/m. 
Modulation ratio $\alpha$ modulates the compliance between PD control and compliance control in the case of $\bm K^{ee}$ = 100 N/m. 

This table allows us to derive a wealth of information as follows. 
First, when comparing the three central rows corresponding to the proposed method under varying task stiffness, it can be observed that increasing stiffness leads to smaller tracking errors in the torso and hand, while simultaneously increasing the torque demands on the arms. 
Moreover, while the basic method yields reduced torso and hand tracking error, it comes at the cost of increased joint energy expenditure (larger upper-body torques). 
For further comparison, we employed a soft joint P gain ($K_p$ = 30). The results indicate that while this setting leads to smaller torso tracking errors (meaning the task-space impact exerts less influence on the body), and lower torque demands, it also produces the largest hand tracking mean error (0.11 m). Hence, simply reducing the joint P gain is impractical in the context of robot loco-manipulation.  
In the final category presented in the table, different ratio $\alpha$ are employed to evaluate the compliance modulation performance of the two control laws.
These results show that the modulation integrates the strengths of both controllers: smaller tracking error compared with compliance control, and lower torque consumption compared with pure PD control. 
In practical robotic applications, the ratio can be tuned to accommodate different task requirements. 

We contend that the limited arm DOFs (4 DOFs per arm) in the humanoid H1 restrict the proposed method from demonstrating its full advantages. When applied to a more redundant robot, such as the humanoid robot G1, these advantages are expected to be more pronounced.

\begin{table}[t]
  \centering
  \begin{threeparttable}
    \caption{Impact Test Results}
    \label{tab:2}
    \scriptsize
    \setlength{\tabcolsep}{3pt}
    \renewcommand{\arraystretch}{1.1}

    \subfloat[Stance Mode\label{table:subtab11}]{%
      \begin{minipage}[t]{.48\linewidth}\centering
      \begin{tabular}{lccc}
        \toprule
        \textbf{Method} & $e^{\text{torso}}$ & $e^{\text{ee}}$ & $J^{\text{upper}}$ \\
        \midrule
        PD (FALCON)              & $0.701 _{\pm 0.221}$ & $0.092 _{\pm 0.055}$ & $49.723$ \\
        Soft P ($K_p$ = 30)      & $0.670 _{\pm 0.192}$ & $0.110 _{\pm 0.055}$ & $43.627$ \\
        \midrule
        $\bm K^{ee}$=100 N/m (CoTaP) & $0.861 _{\pm 0.398}$ & $0.101 _{\pm 0.055}$ & $46.664$ \\
        $\bm K^{ee}$=500 N/m (CoTaP) & $0.759 _{\pm 0.273}$ & $0.095 _{\pm 0.055}$ & $48.648$ \\
        $\bm K^{ee}$=800 N/m (CoTaP) & $0.771 _{\pm 0.274}$ & $0.096 _{\pm 0.055}$ & $50.812$ \\
        \midrule
        $\alpha$=0.3 (CoTaP)     & $0.711 _{\pm 0.224}$ & $0.093 _{\pm 0.055}$ & $48.478$ \\
        $\alpha$=0.7 (CoTaP)     & $0.758 _{\pm 0.261}$ & $0.096 _{\pm 0.055}$ & $47.507$ \\
        \bottomrule
      \end{tabular}
      \end{minipage}
    }\hfill
    \subfloat[Walking Mode\label{table:subtab12}]{%
      \begin{minipage}[t]{.48\linewidth}\centering
      \begin{tabular}{lccc}
        \toprule
        \textbf{Method} & $e^{\text{torso}}$ & $e^{\text{ee}}$ & $J^{\text{upper}}$ \\
        \midrule
        PD (FALCON)              & $2.126 _{\pm 0.734}$ & $0.090 _{\pm 0.055}$ & $50.811$ \\
        Soft P ($K_p$ = 30)      & $2.071 _{\pm 0.720}$ & $0.112 _{\pm 0.055}$ & $44.440$ \\
        \midrule
        $\bm K^{ee}$=100 N/m (CoTaP) & $2.194 _{\pm 0.755}$ & $0.100 _{\pm 0.054}$ & $46.541$ \\
        $\bm K^{ee}$=500 N/m (CoTaP) & $2.153 _{\pm 0.737}$ & $0.094 _{\pm 0.054}$ & $49.169$ \\
        $\bm K^{ee}$=800 N/m (CoTaP) & $2.154 _{\pm 0.727}$ & $0.094 _{\pm 0.054}$ & $51.565$ \\
        \midrule
        $\alpha$=0.3 (CoTaP)     & $2.106 _{\pm 0.716}$ & $0.092 _{\pm 0.057}$ & $48.944$ \\
        $\alpha$=0.7 (CoTaP)     & $2.123 _{\pm 0.740}$ & $0.095 _{\pm 0.054}$ & $47.524$ \\
        \bottomrule
      \end{tabular}
      \end{minipage}
    }

    \begin{tablenotes}
    \centering
      \footnotesize
      \item[*] In the table, the larger values denote the means, and the smaller values following the ± symbol represent the standard deviations.
    \end{tablenotes}
  \end{threeparttable}
\end{table}

\subsubsection{Impact Force at End-effector in Walking Mode}

When the robot is walking at a given velocity, a random external impact is applied, and the average performance across different compliance parameter settings is subsequently evaluated over multiple metrics in \sectref{sec:crit}. 
All the results are presented in Table \ref{table:subtab11}. 
The analysis of the walking state results is largely consistent with the static case, with the main difference being a substantial increase in torso tracking error. 
Meanwhile, incorporating the ratio $\alpha$ for modulated joint-space stiffness highlights more clearly the advantages of reducing tracking error while balancing joint torque consumption.

\figref{fig:walk} illustrates the screenshots of the robot's reaction under an impact during walking in the case of $\alpha$ = 0.7. 
In the figure, the robot is depicted walking forward, with the upper body simultaneously tracking a punching motion. 
After the left hand experiences an external impact (red line in the figure), it is pulled straight; subsequently, due to task-space stiffness, the left hand returns to its original trajectory, while the body, in order to maintain the target velocity after the impact, shows a backward and rotate tendency. 
An external force of this magnitude was not accounted for in the policy training, as the objective of compliance control is not to achieve exact trajectory tracking at every instant in task space, but to preserve compliance and stability in the presence of unforeseen large disturbances. 
As indicated in the title, the primary aim of the proposed method is to realize adjustable task-space compliance. 
Accordingly, these results can be regarded as meeting the requirements of our research objectives.

\begin{figure}[t]
	\begin{center}
	 \includegraphics[width=0.88\hsize]{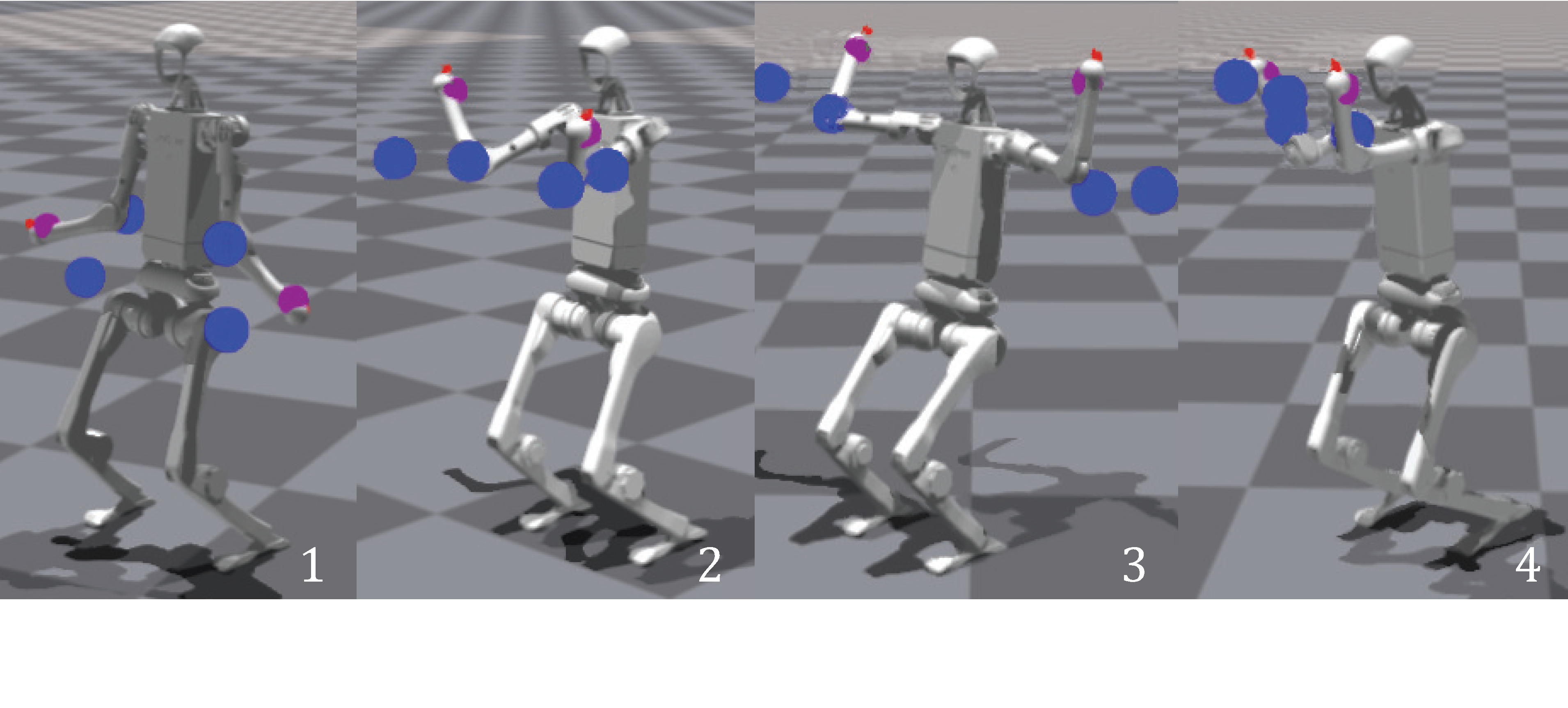}
	 \caption{Policy distillation results without considering Jacobian condition number. At the singular postures illustrated in the figure (e.g., when the arms are fully extended), the compliance control yields unstable joint stiffness. Consequently, the RL training tends to avoid these postures, leading to larger upper-body tracking errors. The blue balls are reference points of hands and elbows. }
    \label{fig:singular}
	\end{center}
\end{figure}

\subsubsection{Ablation Study}

For ablation study, we first tested the compliance controller directly on the base policy without distillation, but effective control could not be achieved, with the robot failing to maintain stability. 
The main issue arose from the lower-body motion failing to adapt to the upper-body control law. 
The results are presented in Table \ref{tab:4}.
Then, for the upper body we directly employed the original reference trajectory as the control target at each timestep. 
Although the control was generally realizable, the errors remained considerable, with the upper-body tracking error (0.252) exceeding that of the proposed method (0.107) by more than twice. 
The reason is still that the upper-body motion was not trained in coordination with the lower-body policy. 
Moreover, to demonstrate the importance of the modulation ratio $\alpha$ considering Jacobian condition number, we conducted an ablation study of training the policy without setting $\alpha$. 
\figref{fig:singular} illustrates a frequent phenomenon observed after training: due to the influence of singular postures, the policy deliberately avoids joint singularity in order to maintain control stability, which in turn leads to larger tracking errors (0.517 for upper-body position).

\begin{table}[t]
\centering
\caption{Ablation Study Result (Stance mode, $\bm K^{ee}$ = 100 N/m) \label{tab:4}}
\begin{tabular}{l|l|l|l}
\hline
\textbf{Method} & \textbf{No-train} & \textbf{w/o} $\alpha$ & \textbf{CoTaP} \\
\hline
$e^{\text{torso}}$ & $0.950 _{\pm 0.621}$ & $1.008 _{\pm 0.370}$ & $0.861 _{\pm 0.398}$ \\
$e^{\text{upper}}$ & $0.252 _{\pm 0.173}$ & $0.517 _{\pm 1.298}$ & $0.107 _{\pm 0.105}$ \\
\hline
\end{tabular}
\end{table}

\subsection{Sim-to-sim Validation}

\begin{figure}[t]
  \centering
  \subfloat[]{%
    \includegraphics[width=0.33\linewidth]{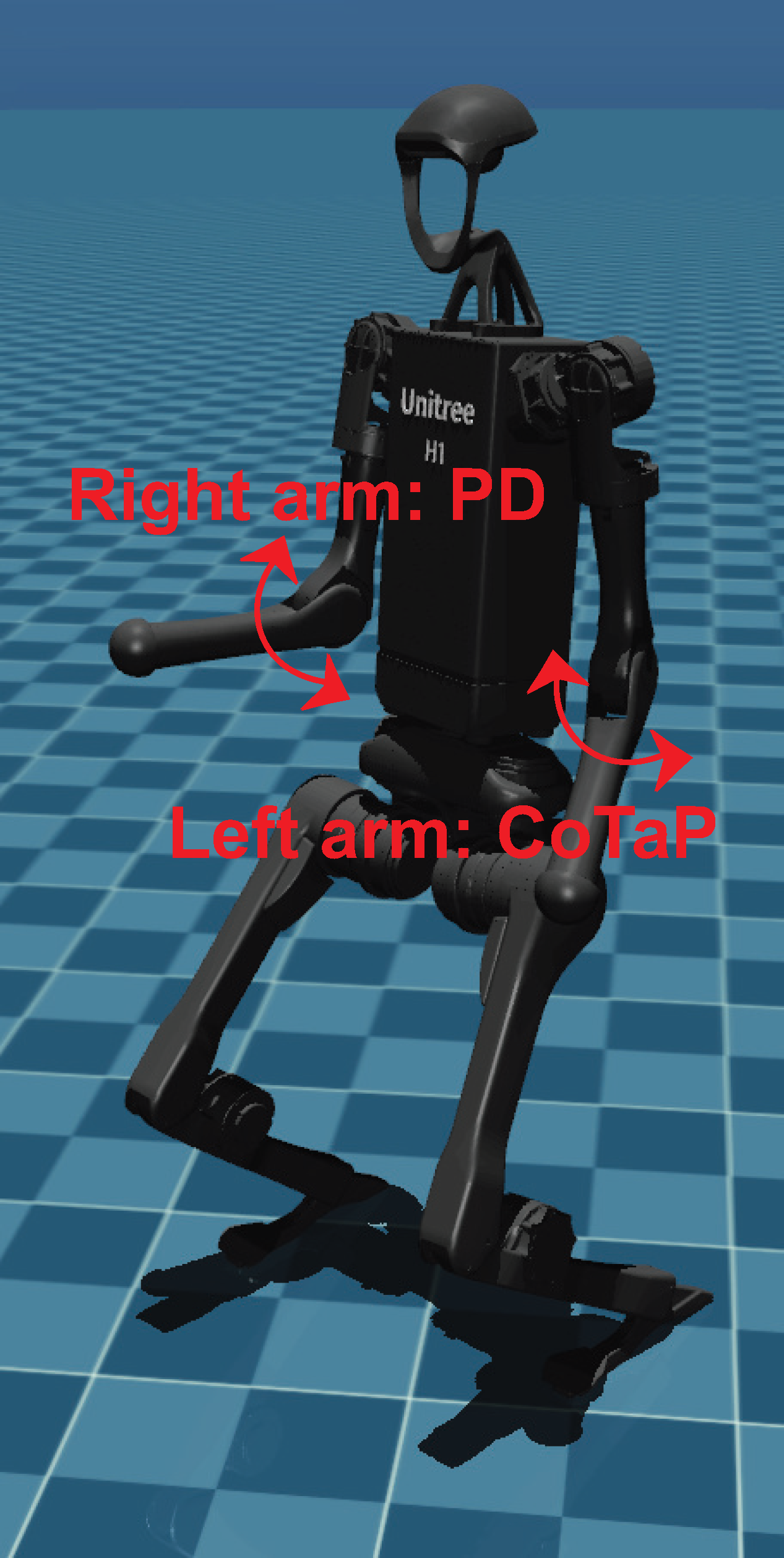}%
    \label{fig07a}
  }\hfill
  \subfloat[]{%
    \includegraphics[width=0.64\linewidth]{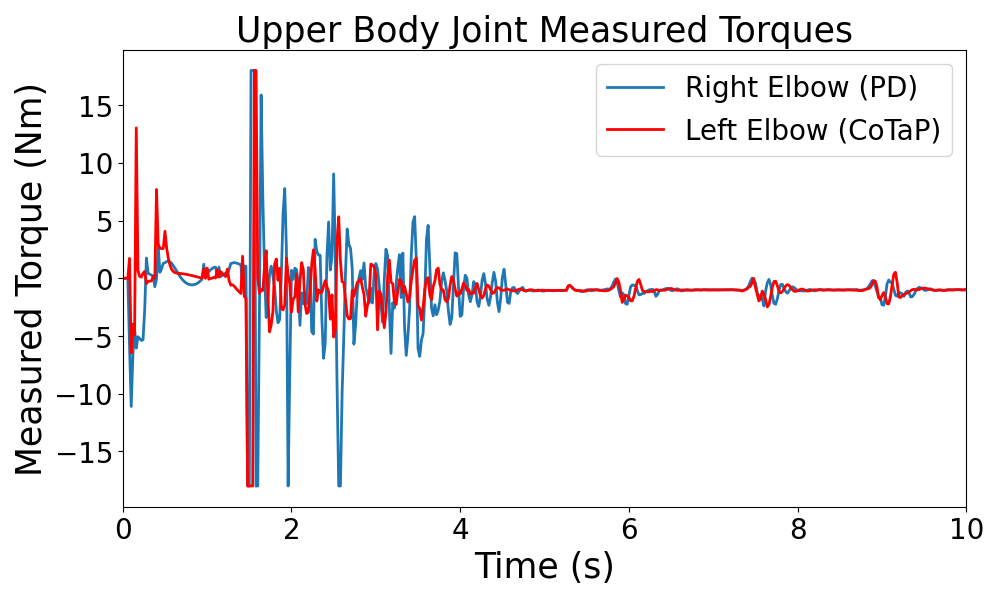}%
    \label{fig07b}
  }
  \caption{(a) Simulation of H1 in MuJoCo under a periodic load applied on both hands. In this simulation, right arm is using PD control and left arm is using modulated compliance control ($\alpha$ = 0.7). The left hand swings with a larger amplitude than the right hand. (b) Torque curves of both elbow joints under periodic loading. It can be seen that the torque of the left arm is overall much smaller than that of the right arm.}
  \label{fig:torque_mjc_sim2sim}
\end{figure}

We applied the policy trained in Isaac Gym together with the overall control method to the MuJoCo simulation \cite{todorov2012mujoco}, thereby achieving sim-to-sim transfer. 
In this simulation, beyond basic velocity-tracking locomotion and upper-body control, a periodic load experiment was performed: with the robot’s arms maintained in their initial configuration, a sinusoidal external force in the $z$-direction, with a period of 4 s and an amplitude of 30 N, was applied to both hands (as shown in \figref{fig07a}). 
Meanwhile, different control methods were applied to the robot’s arms: the right arm was controlled using pure PD, whereas the left arm employed compliance modulation control with $\alpha$ = 0.7 (between PD and $\bm K^{ee}$ = 500 N/m). 
\figref{fig07b} shows the results of the measured elbow torques. 
In the figure, we can observe that the joint torques under modulated compliance control (CoTaP) are much smaller than those under PD control. 
These results demonstrate that, in practical applications, the task-space compliance value together with the modulation ratio $\alpha$ can be varied to quantitatively regulate the robot’s compliance, enabling adaptation to specific objectives such as precision control and energy conservation. 

\section{Conclusions and Future Work}

In this study, we proposed CoTaP, a pipeline applying compliance information in the RL structure, and introduced its controller based on RL and compliance modulation for humanoid robot loco-manipulation. 
The main contribution of this work is the integration of model-based compliance control into the RL training framework, enabling the robot to benefit from the advantages of RL-based motion generation while ensuring quantitatively adjustable compliance at the low-level controller. 
By incorporating randomization of the required task-level compliance values, the controller no longer needs to be retrained each time the compliance setting is modified. 

After achieving this, task-space compliance becomes a parameterizable and adjustable quantity, both for future tele-operation and real-world control. This broadens the overall state space of humanoid robot operation, enhances the adaptability of robotic loco-manipulation, and enables the robot to go beyond simply imitating human motions by compliance modulation according to the actual situation.



\bibliographystyle{IEEEtran}
\bibliography{ral_hrvc}

\clearpage
\appendix

\subsection{Upper-body Tracking Keypoint Setting}

Because the upper and lower body are controlled independently in this study, we do not directly use the keypoint positions (and velocities) in the world frame from the original motion dataset; instead, we first transform them into the torso-link coordinate frame: 
\begin{align}
\bm p^{\text{rel}}_{\text{ee}}
&= \bm R\!\left(\bm q^{\text{ref}}_{\text{torso}}\right)^{-1}
   \big(\widetilde{\bm p}^{\text{ref}}_{\text{ee}}-\bm p^{\text{ref}}_{\text{torso}}\big), \\
\bm v^{\text{rel}}_{\text{ee}}
&= \bm R\!\left(\bm q^{\text{ref}}_{\text{torso}}\right)^{-1}
   \big(\widetilde{\bm v}^{\text{ref}}_{\text{ee}}-\bm v^{\text{ref}}_{\text{torso}}\big)
   \;-\; \boldsymbol\omega^{\text{ref}}_{\text{torso}} \times \bm p^{\text{rel}}_{\text{ee}}.
\end{align}
where $\widetilde{\bm p}^{\text{ref}}_{\text{ee}}$ and $\widetilde{\bm v}^{\text{ref}}_{\text{ee}}$ are the original reference positions and velocities of keypoints. 
Then, we transform them into the world frame again using the current torso-link frame: 
\begin{align}
\bm p^{\text{ref}}_{\text{ee}}
&= \bm R\!\left(\bm q^{\text{cur}}_{\text{torso}}\right)\bm p^{\text{rel}}_{\text{ee}}
   + \bm p^{\text{cur}}_{\text{torso}}, \\
\bm v^{\text{ref}}_{\text{ee}}
&= \bm R\!\left(\bm q^{\text{cur}}_{\text{torso}}\right)
   \Big(\bm v^{\text{rel}}_{\text{ee}}
       + \boldsymbol\omega^{\text{cur}}_{\text{torso}} \times \bm p^{\text{rel}}_{\text{ee}}\Big)
   + \bm v^{\text{cur}}_{\text{torso}}.
\end{align}
where $\bm p^{\text{ref}}_{\text{ee}}$ and $\bm v^{\text{ref}}_{\text{ee}}$ will be used in upper-body keypoints tracking. 


\subsection{Parameter Settings of RL Training \label{sect:app02}}

We applied the similar reward settings from \cite{zhang2025falcon}, but introduce a reference upper-body action close term as explained in \sectref{sect:training}, and a upper-body key point tracking term as follows: 
\begin{align}
    \exp\left(-\frac{1}{0.01} \lVert \bm p_{\text{upper}} - \bm p_{\text{ref}}\rVert_2^2 \right)
\end{align}
Table \ref{tab:obs} shows the observation settings of our method. 
In the current observation space, we reintroduce foot contact phase information while additionally incorporating upper-body joint torque. The details regarding the configuration of upper-body torque term will be elaborated in Appendix \ref{sect:abla_obs}.

\begin{table}[h]
\centering
\caption{Observation Information in Policy setting \label{tab:obs}}
\begin{threeparttable}
\begin{tabular}{|l|c|c|}
\hline
\textbf{State term} & \textbf{Dimensions} & \textbf{Scales} \\
\hline
\multicolumn{3}{|c|}{\textbf{Actor Observation}} \\
\hline
Torso angular velocity & 3 & 0.25 \\
Projected gravity & 3 & 1 \\
Command linear velocity & 2 & 1 \\
Command angular velocity & 1 & 1 \\
Command stand & 1 & 1 \\
Command torso height & 1 & 2 \\
Reference upper-body DOF position & 8 & 1 \\
Whole-body DOF position & 19 & 1 \\
Whole-body DOF velocity & 19 & 0.05 \\
Actions & 19 & 1 \\
\texttt{sin} function of phase time & 1 & 1 \\
\texttt{cos} function of phase time & 1 & 1 \\
Calculated upper-body joint torque \tnote{*} & 8 & 1 \\
\hline
\multicolumn{3}{|c|}{\textbf{Critic Observation} (beyond actor)} \\
\hline
Torso orientation & 4 & 1 \\
Torso linear velocity & 3 & 2 \\
Left EE applied force & 3 & 0.1 \\
Right EE applied force & 3 & 0.1 \\
\hline
\end{tabular}
\begin{tablenotes}
\footnotesize
\item[*] A comparison was conducted between incorporating this component into the actor and into the critic, with the discussion provided in the following section. 
\end{tablenotes}
\end{threeparttable}
\end{table}

\subsection{Ablation Study of Upper-body Joint Torque in Observation \label{sect:abla_obs}}

\begin{figure}[t]
\centering
\subfloat[Actor observation with upper-body torque. ]{
\centering
\includegraphics[width=0.75\hsize]{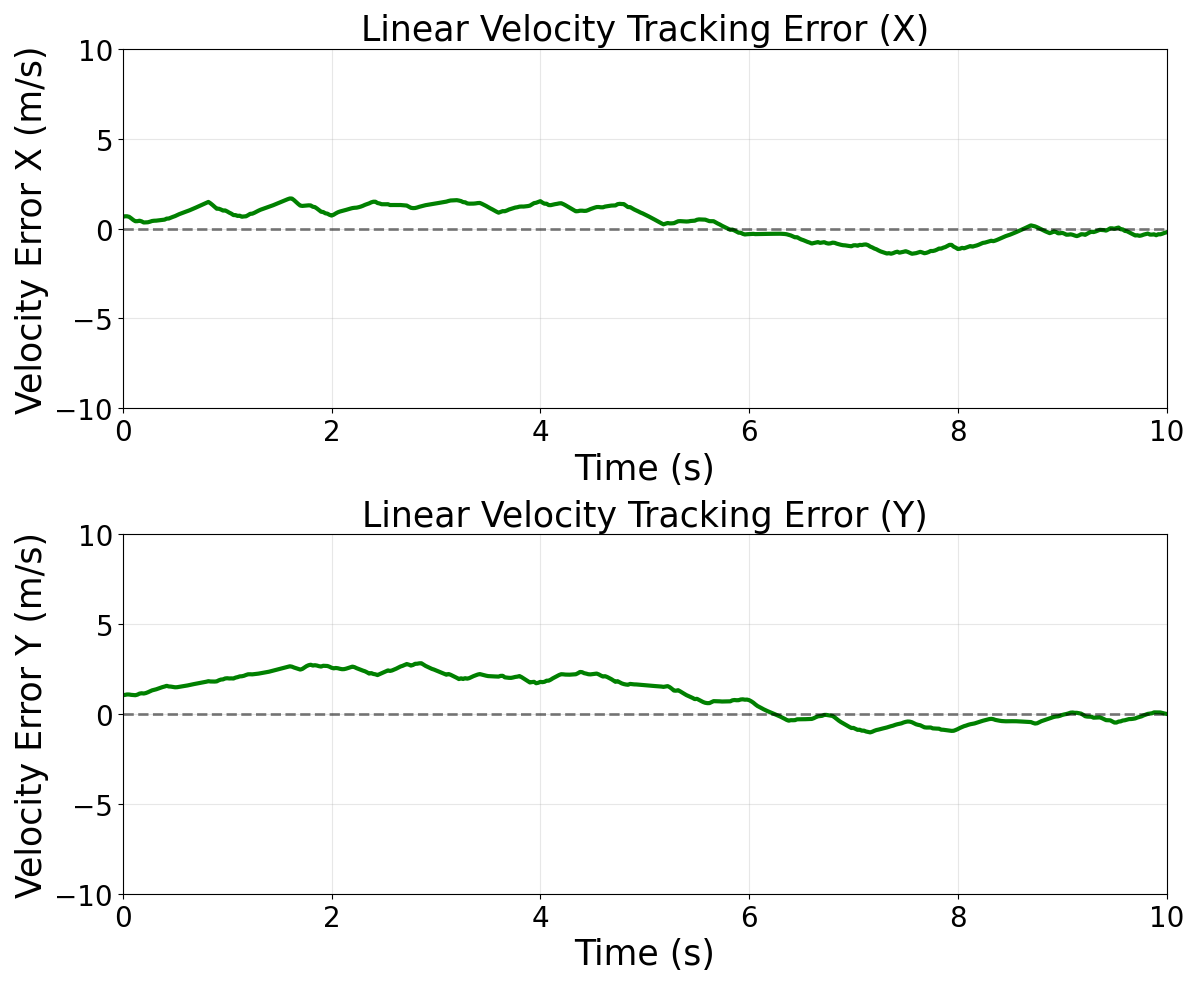}
\label{fig09a}
}%

\subfloat[Actor observation without upper-body torque. ]{
\centering
\includegraphics[width=0.75\hsize]{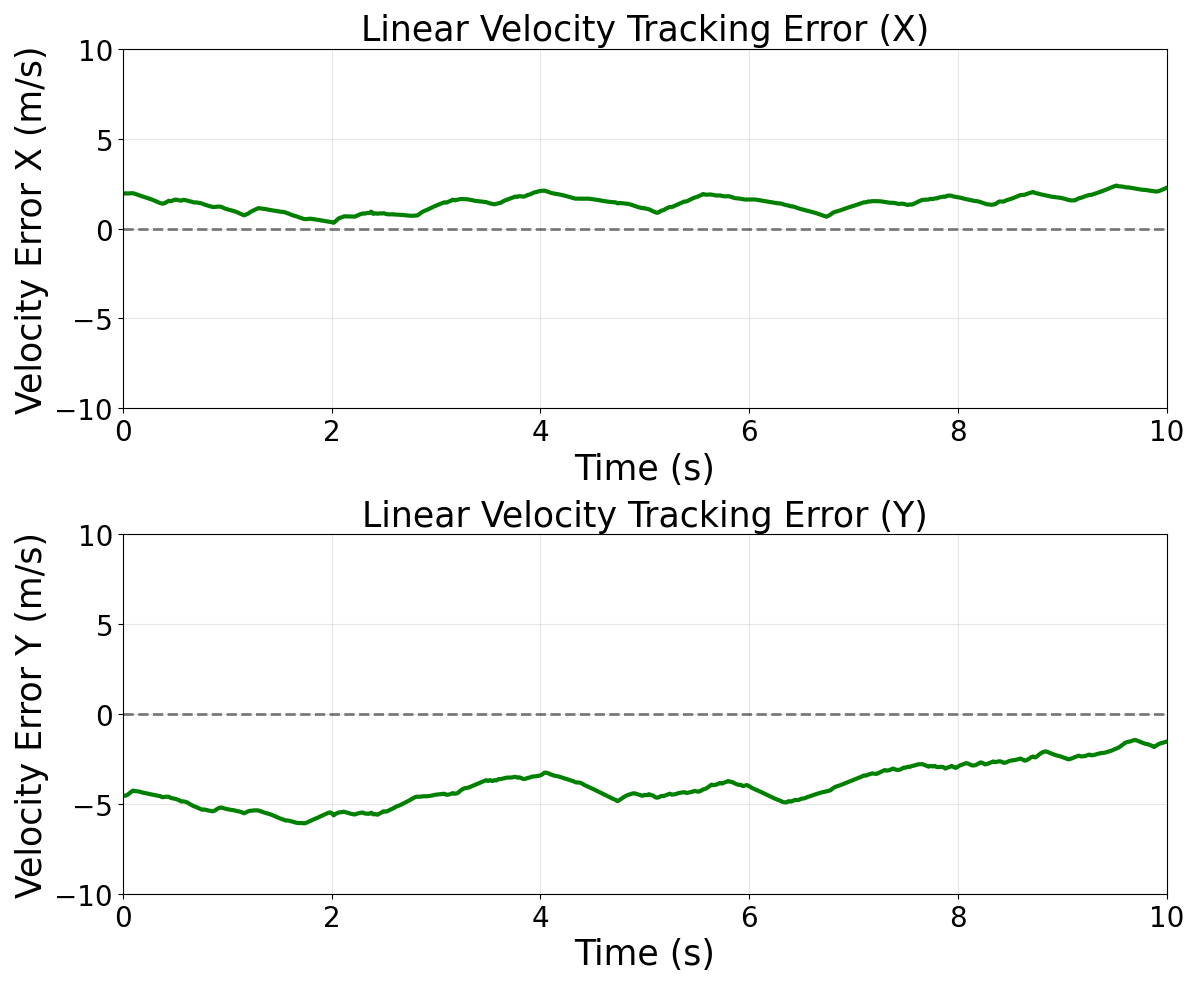}
\label{fig09b}
}
\caption{Linear velocity error in $x$ and $y$ of stepping motion in MuJoCo simulation. \label{fig09}}
\end{figure}

We tested on sim-to-sim in MuJoCo whether to include upper-body joint torque in the actor observations, and the results are shown in \figref{fig09}. 
It can be clearly observed from the results that including upper-body torque within actor observation leads to a notably smaller error compared to the case without it. 
The underlying reason, in our view, is that when upper-body torque is excluded from the actor observation space, the policy has no access to the corresponding control law and parameter settings. 
As a result, while balance can still be maintained to some extent, achieving precise velocity control becomes substantially more challenging.

\subsection{Comparison with FACET method}

According to FACET method \cite{xu2025facet}, we can define a task-space impedance control achieving by position and velocity tracking. 
The definition of impedance force on the end-effector is
\begin{align}
\label{eq:task_compliance}
\bm f_{\text{spring}}
= \bm K_e\bigl(\bm x_{\text{des}} - \bm x\bigr)
+ \bm D_e\bigl(\dot{\bm x}_{\text{des}} - \dot{\bm x}\bigr),
\end{align}
where the $\bm K_e$ and $\bm D_e$ are task-space stiffness and damping matrices. 
Then, the corresponding task-space acceleration term is
\begin{align}
    \ddot{\bm x}_{\rm ref} = \frac{1}{m} \bigl(\bm f_{\text{spring}} + \bm f_{\text{ee}}\bigr), 
\end{align}
where $\bm f_{\text{ee}}$ is the estimated end-effector force; $m$ is the task-space inertia term (in this study we set it as constant virtual mass). In the simulation training, it is critic observation; in the sim-to-real control, it can be measured by force sensor. 
After that, the required task-space velocity and position can be integrated as follows: 
\begin{align}
    \dot{\bm x}_{\rm ref} = \dot{\bm x}(t_0) + \int_{t_0}^{t}\ddot{\bm x}_{\rm ref}(\tau)d\tau, \\
    {\bm x}_{\rm ref} = {\bm x}(t_0) + \int_{t_0}^{t}\dot{\bm x}_{\rm ref}(\tau)d\tau
\end{align}
The required position and velocity will be used in the reward function: 
\begin{align}
    r_{t}
  = \frac{1}{M} \sum_{t'} \exp \Bigl(
      -\bigl\lVert
        \dot{\bm x}_{\rm ref}^{t'} - \dot{\bm x}_{\rm ee}
      \bigr\rVert_{2}^{2}
      -\bigl\lVert
        {\bm x}_{\rm ref}^{t'} - {\bm x}_{\rm ee}
      \bigr\rVert_{2}^{2}
    \Bigr)
\end{align}
This term is used to replace upper-body key point tracking. 
However, since directly introducing the ${\bm x}_{\rm ref}$ and $\dot{\bm x}_{\rm ref}$ terms during training requires modifying the upper-body trajectories through IK, we only incorporate position and velocity tracking based on the original dataset, without online integration used in FACET. 
The acceleration–velocity–position integration is applied during the evaluation stage, where real-time IK is also introduced. 

As the result, we made a comparison between \textit{only CoTaP}, \textit{only FACET}, and \textit{CoTaP+FACET} as in Table \ref{tab:apx}. 
This table shows the end-effector static error in $z$-direction under different loads. The end-effector stiffness is set as 500 N/m. 
The \textit{Ideal} column shows the reference error calculated by given load and stiffness. 
We observe that with only CoTaP, the error relative to the ideal is large, even reversing sign at low loads; by contrast, only FACET and CoTaP+FACET, while still deviating from the ideal, are relatively more accurate.
This finding indicates that the FACET method plays a critical role in reducing steady-state error.
This is intuitive: the FACET method explicitly computes and tracks the required task-space displacement. In the steady state, as long as the RL-trained policy tracks well, static task-space compliance can be achieved with high precision.

\begin{figure}[t]
\centering
\subfloat[only FACET]{
\centering
\includegraphics[width=0.75\hsize]{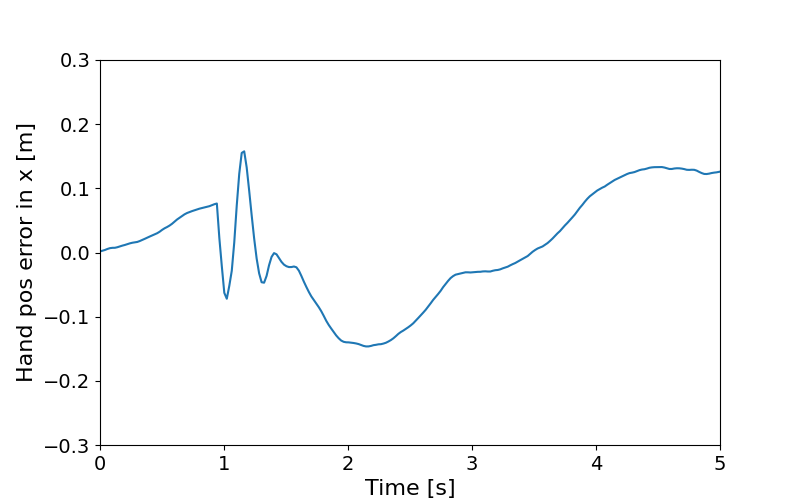}
\label{fig09a}
}%

\subfloat[only CoTaP]{
\centering
\includegraphics[width=0.75\hsize]{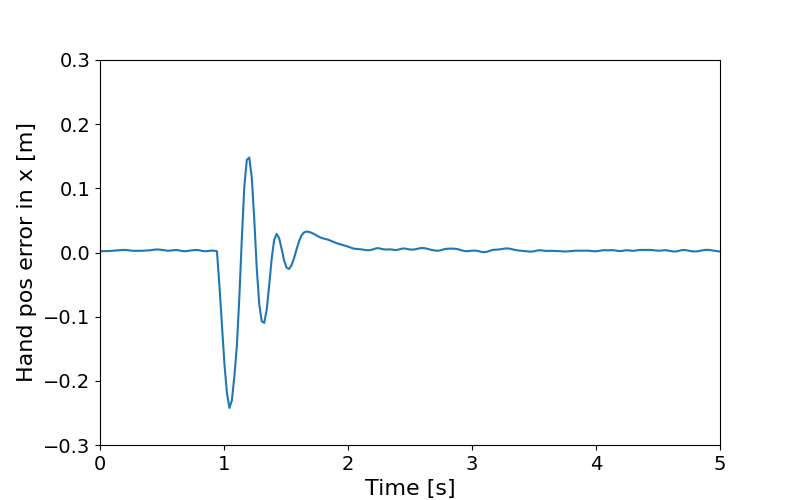}
\label{fig09b}
}
\caption{Hand position error in $x$ after external impact (500 N in 0.05 s at 1 s). \label{fig10}}
\end{figure}

On the other hand, we designed a transient-impact experiment to compare the methods above. 
Similar as in \sectref{sect:simulation}, we added an external impact on robot's left hand in stance mode. 
The end-effector stiffness of both methods are set as 500 N/m. 
\figref{fig10} shows the hand position error curves in only FACET and only CoTaP simulations, separately. 
From the figures, we can see that only CoTaP method achieves a better post-impact response than only FACET method. 
In the results, we can find that under transient impacts the high-frequency oscillatory behavior with the FACET method remains dominated by the joint-level PD controller, which is completely different from the CoTaP method.
Furthermore, because the FACET method relies on online IK computation, its control of the end-effector position during dynamic responses is suboptimal.

\begin{table}[t]
\centering
\caption{Comparison of end-effector position errors in $z$ under different load magnitudes (unit: m; end-effector stiffness: 500 N/m) \label{tab:apx}}
\begin{tabular}{c|l|l|l|l}
\hline
\textbf{Load} & \textbf{Ideal} & \textbf{only CoTaP} & \textbf{only FACET} & \textbf{CoTaP+FACET} \\
\hline
$10 N$ & $0.02$ & $-0.003$ & $0.043$ & $0.032$ \\
$30 N$ & $0.06$ & $-0.015$ & $0.068$ & $0.046$ \\
$50 N$ & $0.10$ & $0.058$ & $0.106$ & $0.097$ \\
\hline
\end{tabular}
\end{table}

Overall, CoTaP and FACET methods each have strengths and weaknesses in controlling steady-state and transient errors. 
This is consistent with the underlying theoretical derivations: we view the two methods as regulating compliance at two distinct levels. 
How to comprehensively analyze and control these two forms of compliance in practical implementations will be a focus of our future work.




\end{document}